\documentclass[journal]{IEEEtran}
\usepackage{pgfplots}
\usepackage{tikz}
\usetikzlibrary{shapes.geometric, arrows}
\usetikzlibrary{patterns}
\pgfplotsset{compat=1.18}
\usepackage{amsmath}
\usepackage[dvipsnames]{xcolor}
\usepackage[hidelinks]{hyperref}
\usepackage{multirow}
\usepackage{nicematrix}
\usepackage{flushend}

\usepackage{glossaries}

\newacronym{ml}{ML}{machine learning}
\newacronym{dl}{DL}{deep learning}
\newacronym{iot}{IoT}{Internet of Things}
\newacronym{mcu}{MCU}{microcontroller unit}
\newacronym{cnn}{CNN}{convolutional neural network}
\newacronym{dnn}{DNN}{deep neural network}
\newacronym{vit}{ViT}{vision transformer}
\newacronym{map}{mAP}{mean Average Precision}
\newacronym{soc}{SoC}{system-on-chip}
\newacronym{ai}{AI}{artificial intelligence}
\newacronym{mac}{MAC}{multiply-accumulate}
\newacronym{yolo}{YOLO}{You Only Live Once}
\newacronym{mse}{MSE}{mean squared error}
\newacronym{iou}{IoU}{intersection over union}
\newacronym{c2f}{C2f}{cross-stage partial bottleneck with two convolutions}
\newacronym{sppf}{SPPF}{spatial pyramid pooling - fast}
\newacronym{ptq}{PTQ}{post training quantization}
\newacronym{fps}{fps}{frames per second}
\newacronym{lsi}{LSI}{Low-Speed Internal oscillator}
\newacronym{msi}{MSI}{Multi-Speed Internal oscillator}
\newacronym{rtc}{RTC}{Real-Time Clock}

\ifCLASSINFOpdf
\else
\fi
\begin{document}
%
\title{Hardware-Aware YOLO Compression for Low-Power Edge AI on STM32U5 for Weeds Detection in Digital Agriculture}
%
%
%

\author{Charalampos~S.~Kouzinopoulos
        and~Yuri~Manna
\thanks{The authors are with the Department of Advanced Computing Sciences, Maastricht University, 6229 GS, Maastricht, Netherlands. e-mail: charis.kouzinopoulos@maastrichtuniversity.nl, y.manna@student.maastrichtuniversity.nl (\textit{Corresponding author: C.S. Kouzinopoulos}).}
\thanks{Manuscript received April 19, 2005; revised August 26, 2015.}}

%
%

\markboth{Journal of \LaTeX\ Class Files,~Vol.~14, No.~8, August~2015}%
{Shell \MakeLowercase{\textit{et al.}}: Bare Demo of IEEEtran.cls for IEEE Journals}
%



\maketitle

\begin{abstract}
Weeds significantly reduce crop yields worldwide and pose major challenges to sustainable agriculture. Traditional weed management methods, primarily relying on chemical herbicides, risk environmental contamination and lead to the emergence of herbicide-resistant species. Precision weeding, leveraging computer vision and machine learning methods, offers a promising eco-friendly alternative but is often limited by reliance on high-power computational platforms. This work presents an optimized, low-power edge AI system for weeds detection based on the YOLOv8n object detector deployed on the STM32U575ZI microcontroller. Several compression techniques are applied to the detection model, including structured pruning, integer quantization and input image resolution scaling in order to meet strict hardware constraints. The model is trained and evaluated on the CropAndWeed dataset with 74 plant species, achieving a balanced trade-off between detection accuracy and efficiency. Our system supports real-time, in-situ weeds detection with a minimal energy consumption of 51.8mJ per inference, enabling scalable deployment in power-constrained agricultural environments.
\end{abstract}

\begin{IEEEkeywords}
Edge Artificial Intelligence (Edge AI), Computer Vision, Deep Neural Networks (DNNs), Digital Agriculture, Tiny Machine Learning (TinyML), Internet of Things (IoT).
\end{IEEEkeywords}

%
\IEEEpeerreviewmaketitle

\section{Introduction}
%
%
%
%

\IEEEPARstart{W}{eeds} are widespread and persistent plants, known for their rapid reproduction and effective seed dispersal strategies. They are among the primary contributors to crop yield loss globally, posing a significant challenge for farmers and agricultural stakeholders~\cite{junaid2024global}.
Effective weeds management remains a critical issue, as weeds impact broader concerns related to food security, ecosystem health, and environmental sustainability~\cite{vasileiou2024transforming}.
Conventional weeds control strategies have traditionally relied heavily on chemical herbicides. While these methods can be effective in the short term, with a $52$-$96\%$ efficacy over manual weeding~\cite{das2024herbicides}, their widespread and indiscriminate use risks contaminating ecosystems and accelerates the emergence of herbicide-resistant weed species. As of $2024$, more than $523$ unique cases of weed resistance have been reported in $269$ weed species~\cite{das2024herbicides}. 
Addressing this issue aligns directly with the European Green Deal’s objectives of promoting sustainable food systems, reducing chemical pesticide use, and safeguarding biodiversity, highlighting the need for innovative, eco-friendly weed control strategies~\cite{GreenDeal2025}. 

Effective weed control can increase the productivity per unit area to meet the growing demand for crop production~\cite{oerke2004safeguarding}. One promising solution that can be cost-effective and environmentally sustainable is to automate the weeding process through precision weeding. This technique involves the targeted identification and removal of weeds using advanced technologies such as computer vision, \gls*{ml}, robotics and remote sensing. Unlike conventional blanket spraying methods, precision weeding systems can accurately distinguish between crops and weeds, enabling site-specific interventions that minimize chemical usage and reduce damage to surrounding crops and soil. 

A typical weed detection system follows four key steps: image acquisition, data segmentation, feature extraction, and weed detection~\cite{shanmugam2020automated}.
A novel and effective approach for the detection of weeds in image data leverages \gls*{ml} techniques, particularly through the use of computer vision and \gls*{dl} models. There is significant research interest in this area, with several studies demonstrating high accuracy in weed identification using \glspl*{cnn}, \glspl*{vit} and other architectures~\cite{bah2018deep}\cite{xie2021toward}\cite{yu2019deep}. A thorough review on the use of different \gls*{dl} methods for weed detection is given in~\cite{qu2024deep}.
However, most existing work is based on high-end computational platforms for inference, such as GPUs or cloud-based systems. These setups are not directly transferable to real-world agricultural fields, where energy efficiency, latency, cost and connectivity constraints pose significant challenges. 
Many rural areas lack the connectivity required to support sensor node installations, and while power may be sporadically available, it is often scarce across the broader farm landscape.
This highlights the need for the design and deployment of efficient weed detection models on low-power, edge-based platforms, enabling real-time, in-situ decision-making.

\begin{figure}[h]
    \centering
    \includegraphics[width=1\linewidth]{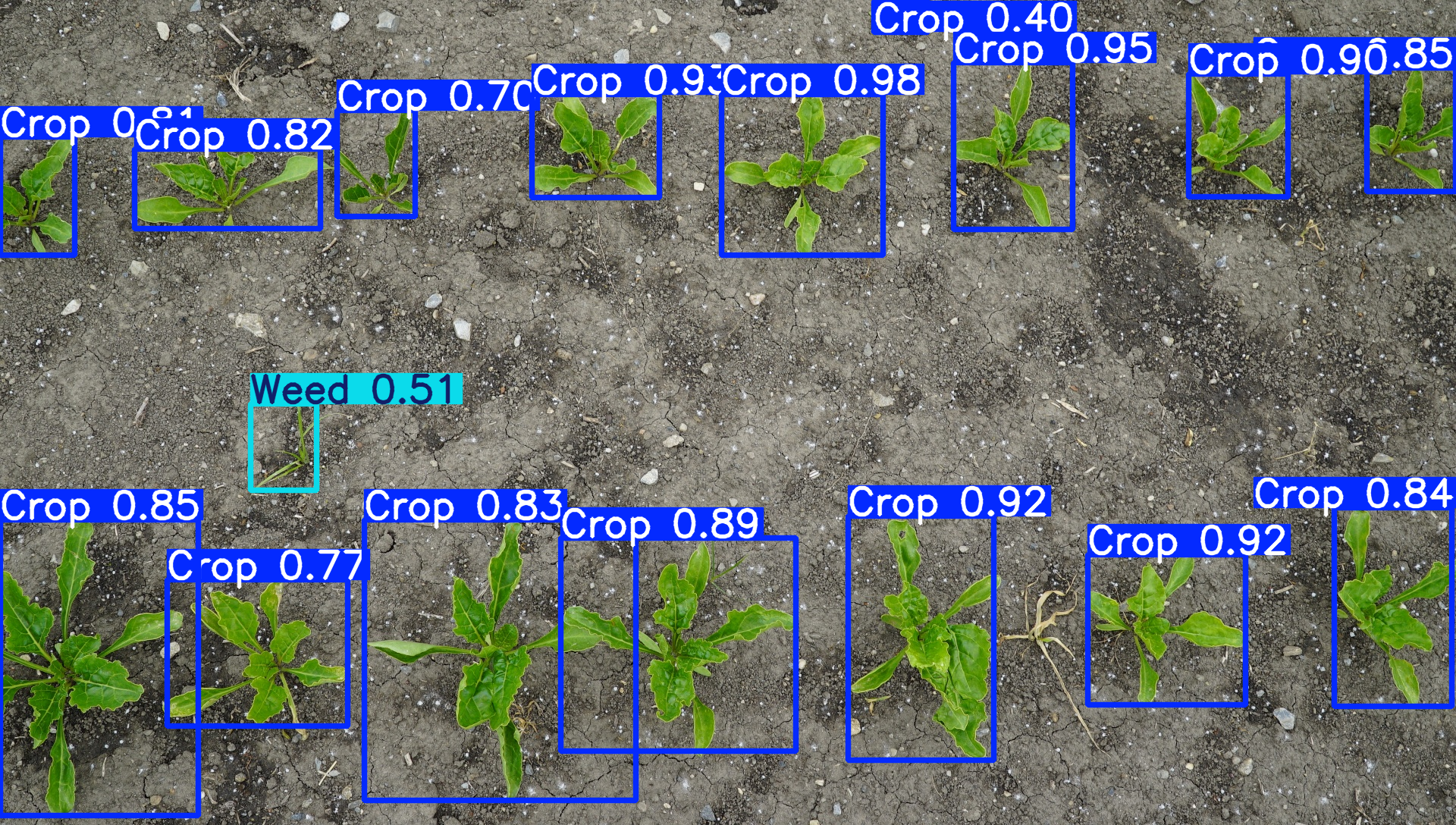}
    \caption{YOLOv8n detection output}
    \label{model_output}
\end{figure}

The aim of this study is to develop a low-power \gls*{iot} system, capable of detecting the presence of weeds in open-air fields and greenhouses, via the use of an optimized computer vision inference pipeline. The proposed pipeline is based on the YOLOv8n one-shot detector, and is executed at the edge targeting the low-power STM32U575ZI \gls*{mcu} by ST Microelectronics.
The key contributions of this article are as follows: 
1) the development of an optimized computer vision edge AI model to enable efficient on-device execution, via the use of different model compression techniques, including structured pruning, integer quantization and input data size reduction; 
2) the evaluation of the model's efficiency vs size of weights and activations on a large-scale dataset; and 
3) the evaluation of the system's energy consumption during edge inference.
Figure~\ref{model_output} shows an example of a successful detection output.

The rest of the paper is organized as follows. 
Section~\ref{sec:related} reviews related work on YOLO-based edge AI methods for low-power object detection as well as on different YOLO architectures for weed detection.
Section~\ref{sec:methodology} details the background and proposed methodology, including training process, pruning and quantization compression techniques as well as the hardware setup and deployment process on the STM32U575ZI \gls*{mcu}.
Section~\ref{sec:evaluation} evaluates the results in terms of detection accuracy, inference time, and energy consumption.
Section~\ref{sec:conclusions} draws conclusions and outlines directions for future research.

\section{Literature Review}
\label{sec:related}

\subsection{Edge AI YOLO networks}
The feasibility of object detection on
resource-constrained, low-power \glspl*{mcu} is an active topic in the literature, driven by the need to bring intelligent capabilities closer to the data source and provide localized decision making. This paradigm, known as \gls*{ai} at the edge or \textit{edge AI}, involves performing inference of \glspl*{dnn} and other \gls{ai} models directly on edge devices, without the need to offload computation to cloud servers or more powerful, energy-intense hardware.
There is a significant body of work on low-power edge AI based on \gls*{cnn} models and single-shot detectors. Specifically, focus is given on the compression of different YOLO (You Only Look Once) versions, a popular single-stage detector that uses \glspl*{cnn} for object detection and classification, for resource-constrained \glspl*{mcu}.

TinyissimoYOLO, a lightweight object detection network was presented in~\cite{moosmann2023tinyissimoyolo}. The network architecture, consisting of $422$k parameters, was compressed using quantization and quantization-aware training techniques enabling deployment on different \glspl*{mcu}.
On the ST Microelectronics STM32H7A3 and STM32L4R9 \glspl*{mcu} it achieved latencies of $359$ms and $996$ms and an energy consumption per inference of $41.8$mJ and $102$mJ respectively.
When deployed on the Ambiq Apollo4b, the network had a latency of $540$ms and $6.08$mJ energy per inference.
Finally, it was deployed on an Analog Devices MAX78000 \gls*{mcu} with a \gls*{cnn} accelerator, achieving a latency of $5.5$ms with an energy consumption of $0.19$mJ per inference.


A modified version of TinyissimoYOLO was introduced in~\cite{boyle2024dsort} and was evaluated on the Greenwaves GAP9 \gls*{mcu} for object detection,  achieving a latency of up to $16.2$ms and energy consumption of up to $31\mu $J per inference. 
Squeezed Edge YOLO, a compressed object detection model with $931$k parameters for human and shape detection was proposed in~\cite{humes2023squeezed}. The model was evaluated on a Greenwaves GAP8 \gls*{mcu}, achieving approximately $70.3$mJ energy ($541$mW power) consumption  and $130$ms latency per inference.

$\mu$YOLO, a custom YOLO architecture for \glspl*{mcu}, was introduced in~\cite{deutel2024microyolo}. The model was deployed on an OpenMV Cam H7 R2 \gls*{mcu}, based on an Cortex-M7 core, with memory requirements of $800$ KB of flash and $350$ KB of RAM. To reduce model size and computational cost, the network was pruned across all convolutional and fully connected layers during training using an iterative gradual pruning schedule guided by an $L1$-norm heuristic, and was further compressed using $8$-bit quantization. Energy consumption on the board and latency are not reported; moreover the size of parameters is not given, although it can be estimated based on the flash requirements and quantization bits to approximately $800$k.

MCUBench, a large-scale benchmark of $100$+ YOLO-based object detection models was presented in~\cite{sah2024mcubenchbenchmarktinyobject}. The models were evaluated on the Pascal VOC dataset~\cite{everingham2015pascal} covering $20$ different object classes. Performance metrics, such as average precision, latency, RAM, and flash usage were reported across various \gls*{mcu} families, including STM32H7, STM32F4, STM32F7, STM32L4 and STM32H5-U5. 

\subsection{YOLO-based weeds detection}
Several YOLO architectures have also been implemented in the literature for the detection of weeds, although not specifically targeting low-power embedded systems.
In~\cite{wang2025lightweight}, YOLO-Weed Nano was presented, a lightweight detection network based on YOLOv8n for weed recognition in cotton fields. The network had a \gls*{map} at an \gls*{iou} threshold of $50$ (mAP50) of $0.931$, $1.09$M parameters and $2.4$MB weight size, achieving $4.7$ GFLOPS and $227.3$ \gls*{fps}. An NVIDIA RTX 3090 GPU was used for the network evaluation. A lightweight YOLO-based model was introduced in~\cite{khater2025ecoweednet} for weed detection, achieving an mAP50 of $0.952$, a \gls*{map} at \gls*{iou} thresholds from $0.50$ to $0.95$ (mAP50-95) of $0.889$ with approximately $2.78$M parameters and $9.3$ GFLOPS. The model was evaluated on an NVIDIA RTX 3080 Ti GPU. YOLO-WDNet, a lightweight model based in YOLOv5 for weed detection in cotton fields was presented in~\cite{fan2024yolo}. The model, evaluated on an NVIDIA Jetson AGX Xavier GPU, had an mAP50 of $0.978$, with $0.9$M parameters and $3.1$ GFLOPS.


To the best of our knowledge, this is the first study in the literature to explore compression techniques of YOLO models for weed detection specifically on low-power \glspl*{mcu}, such as STM32U575ZI, enabling localized, low-power decision-making for the identification of weeds in digital agriculture.

\section{Methodology}
\label{sec:methodology}

\subsection{YOLO (You Only Look Once)}
Traditional \gls*{cnn} object detectors typically follow a two-stage pipeline: first, the input image is scanned to generate region proposals; each region is then classified separately.
In contrast, YOLO is a single-stage detector that processes the entire image in a single pass through convolutional layers, framing object detection as a regression problem. This approach improves significantly computationally efficiency, making YOLO suitable for real-time execution in resource-constrained environments, while still achieving state-of-the-art accuracy.

For this work, three recent versions of YOLO have been initially evaluated: YOLOv8n, YOLOv10n and YOLOv11n. As detailed in section~\ref{sec:pruning} below, YOLOv8n experienced significantly less performance degradation after pruning compared to the YOLO versions evaluated and for this reason was selected as the baseline model for this study. The architecture comprises approximately $3.2 \times 10^6$ parameters distributed across $129$ layers for a total size of $9.95$ MB.
%


YOLOv8 follows an anchor-free detection approach with different stages, where each stage typically consists of multiple convolutional blocks, producing features at a higher level of abstraction and processing data from different resolutions~\cite{yoloarchitecture}. The model architecture is structured around three main components: the backbone for feature extraction at multiple scales, the neck for feature aggregation and the head for prediction. YOLOv8 uses a similar backbone to YOLOv5, starting with the \textit{stem}, where two convolutional blocks convert the input into initial features and reduce its resolution. A series of \Gls*{c2f} blocks follow, incorporating two parallel gradient flow branches to enhance the integration of features with contextual information. They split the input feature map along the channel dimension with one part passing unchanged through an \textit{identity connection} while the other passing through a series of lightweight bottleneck modules.
A \Gls*{sppf} block 
is used to provide a multi-scale representation of the input feature maps. By pooling at different scales, \Gls*{sppf} allows the model to capture features at various levels of abstraction.
The neck assembles feature pyramids, aggregating features from different backbone stages using up-sampling and concatenation, allowing multi-scale context capture.
The head generates the final detection and classification outputs and operates on the fused feature maps to regress bounding box coordinates, confidence scores, and class labels. 


\subsection{Training}
The model was trained on CropAndWeed ~\cite{Dataset}, a large-scale image dataset for the fine-grained identification of crop and weed species. The dataset is composed of annotated samples from cultivation areas in Austria, as part of an \textit{application} set, and additionally includes an \textit{experimental} set with different species, specifically grown in controlled outdoor environments to enrich the dataset with underrepresented classes.
It contains a collection of $74$ plant species: $16$ crop species, including maize, sugar, beets, beans, peas, sunflowers, soy, potatoes and pumpkins, and $58$ weed species, such as grass, amaranth and goosefoot. The images cover varying conditions in terms of lighting conditions as well as soil and type.

For the model training of this work, data were used from both the \textit{application} and \textit{experimental} sets. More than $20.765$ crops instances and  $57.523$ weed instances were represented. The dataset was split into training and validation sets using an $80$-$20$ ratio.
Training of the baseline model was carried out using the Ultralytics library on a NVIDIA GeForce RTX 3090 GPU using CUDA 12.2. The model was trained for $300$ epochs with an input resolution of $224 \times 224$ pixels. All training and fine-tuning hyper-parameters used are listed in Table~\ref{tab1}.


\begin{table}
\centering
\caption{Training and fine-tuning hyperparameters}\label{tab1}
\begin{tabular}{|l|cc|}
\hline
\textbf{Hyperparameter} &  \textbf{Training} & \textbf{Fine-tuning}\\
\hline
Batch size             & Auto (-$1$)     & $150$ \\
Epochs                 & $300$               & $30$ \\
Image size &  \multicolumn{2}{|c|}{$224 \times 224$}\\
Initial learning rate ($lr_0$) & \multicolumn{2}{|c|}{$0.001$}\\
Final learning rate ($lrf$)      & \multicolumn{2}{|c|}{$0.1$}\\
Patience & $50$                & -- \\
Momentum               & $0.937$             & -- \\
Optimizer              & SGD               & Adam \\
Weight decay           & $0.0005 $           & -- \\
Automatic Mixed Precision  & --                & True \\
Test-time augmentation           & \multicolumn{2}{|c|}{True}\\
\hline
\end{tabular}
\end{table}


\subsection{Model Compression}
\label{sec:model_compression}
A major challenge in deploying \glspl*{dnn} on low-power \glspl*{mcu} is meeting the strict constraints on memory, processing power and energy consumption. For the target hardware platform in this work, the model including weights must be compressed to under $2$ MB to fit within the internal flash memory, while the model activations including input and output must fit within $768$ KB of RAM. At the same time, it is essential to minimize the model’s energy consumption without significantly impacting detection accuracy. To address these constraints, we applied a combination of structured pruning and integer quantization. Moreover, the input data dimensions were scaled to $224 \times 224$ pixels to reduce computational load and memory usage. 

\subsubsection{Pruning}
\label{sec:pruning}
According to the Lottery Ticket Hypothesis~\cite{frankle2018lottery}, a randomly initialized network contains sub-networks that, when trained in isolation, can match the performance of the original network. Pruning exploits this concept by eliminating parameters, such as weights and activations, to reduce model size while maintaining target accuracy. It is important to note that filters in different network layers have varying importance to the model’s performance~\cite{wang2024rl}, while aggressive pruning, especially on critical layers such as attention modules, may significantly reduce performance. Pruning is often followed by fine-tuning to improve network accuracy.

\begin{figure}[h]
  \centering
  \begin{tikzpicture}[scale=1]
    \begin{axis}[
        /pgf/number format/1000 sep={},
    width=7.7cm,
    height=3.3cm,
        scale only axis,
        separate axis lines,
        axis on top,
        xmin=0,
        xmax=13,
        xtick={1, 2, 3, 4, 5, 6, 7, 8, 9, 10, 11, 12},
        x tick style={draw=none},
        xticklabels={1, 2, 3, 4, 5, 6, 7, 8, 9, 10, 11, 12},
        xlabel={Pruning steps},
        label style={font=\scriptsize},
        tick label style={font=\scriptsize},
        ymin=0,
        ymax=0.45,
        ylabel={mAP50-95},
        every axis plot/.append style={
          ybar,
          bar width=10pt,
          fill
        }
      ]

      \addplot[draw=black, fill=blue!60!white, postaction={pattern=north east lines}] coordinates{(1, 0.21269778)};
      \addplot[draw=black, fill=blue!60!white, postaction={pattern=north east lines}] coordinates{(2, 0.252299831)};
      \addplot[draw=black, fill=blue!60!white, postaction={pattern=north east lines}] coordinates{(3, 0.28470821)};
      \addplot[draw=black, fill=blue!60!white, postaction={pattern=north east lines}] coordinates{(4, 0.289458281)};
      \addplot[draw=black, fill=blue!60!white, postaction={pattern=north east lines}] coordinates{(5, 0.280961842)};
      \addplot[draw=black, fill=red!70!white, postaction={pattern=north east lines}] coordinates{(6, 0.312297952)};
      \addplot[draw=black, fill=blue!60!white, postaction={pattern=north east lines}] coordinates{(7, 0.30739727)};
      \addplot[draw=black, fill=blue!60!white, postaction={pattern=north east lines}] coordinates{(8, 0.309032973)};
      \addplot[draw=black, fill=blue!60!white, postaction={pattern=north east lines}] coordinates{(9, 0.236364197)};
      \addplot[draw=black, fill=blue!60!white, postaction={pattern=north east lines}] coordinates{(10, 0.241169391)};
      \addplot[draw=black, fill=blue!60!white, postaction={pattern=north east lines}] coordinates{(11, 0.216945615)};
      \addplot[draw=black, fill=blue!60!white, postaction={pattern=north east lines}] coordinates{(12, 0.208960251)};
    
    \end{axis}
  \end{tikzpicture}
  \caption{Effect of $k$ on model accuracy for a $70\%$ pruning ratio}
\label{pruning_steps}
\end{figure}
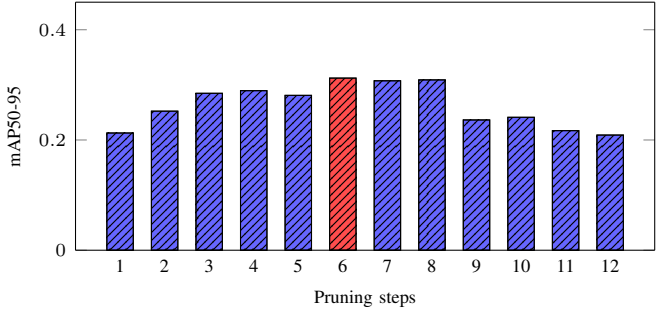

There are two main approaches to network pruning, \textit{unstructured} or \textit{weight pruning} and \textit{structured} or \textit{channel pruning}. In \textit{unstructured} pruning individual weights are removed in a fine-grained way, resulting in a sparse model. However, this often requires specialized hardware to benefit from the sparsity. \textit{Structured} pruning on the other hand involves the removal of entire neurons or filters, preserving the original connectivity and maintaining the dense structure of the network, so that further optimizations can be easily implemented without any specialized hardware units~\cite{kokhazadeh2024denseflex}.
In this work, pruning is applied to entire convolutional filters and channels, since that way a considerable pruning ratio can usually be achieved with little performance degradation~\cite{wang2021convolutional}.

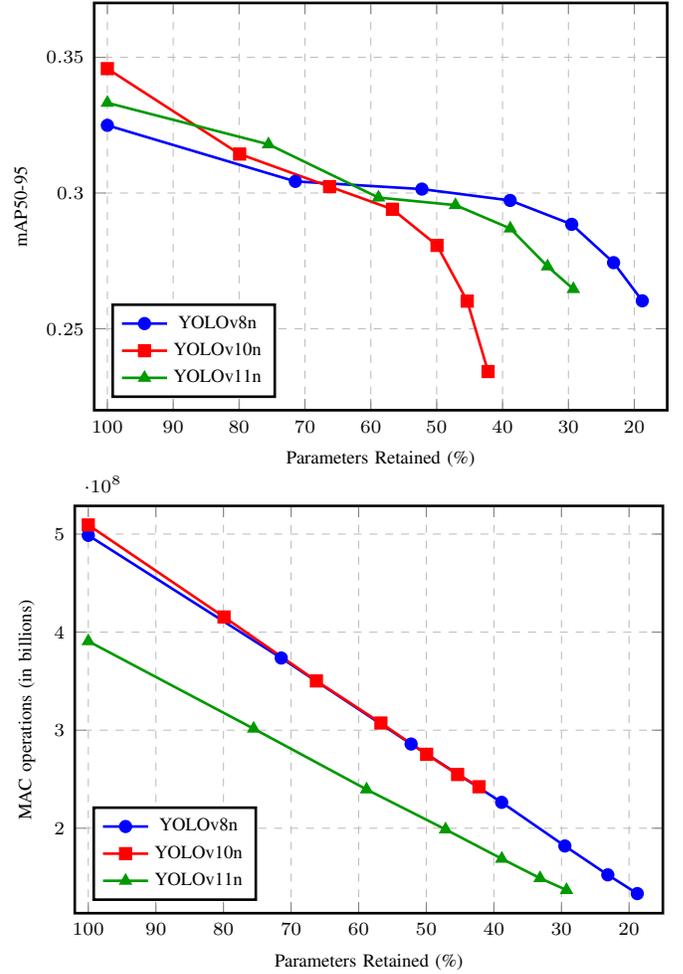
\begin{figure}[h]
\centering
\begin{tikzpicture}
\begin{axis}[
    width=9.2cm,
    height=7cm,
    xlabel={Parameters Retained (\%)},
    ylabel={mAP50-95},
    xmin=15, xmax=102,
    ymin=0.22, ymax=0.37,
    x dir=reverse,
    grid=major,
    grid style=dashed,
    legend pos=south west,
    legend style={font=\scriptsize},
    line width=1pt,
    tick label style={font=\scriptsize},
    label style={font=\scriptsize},
    enlargelimits=false,
]

\addplot[color=blue, mark=*] coordinates {
    (100, 0.3249374630875884)
    (71.45755333852722, 0.3042847672381793)
    (52.245654445115264, 0.3014512809242559)
    (38.852159809354156, 0.29727730451214335)
    (29.502516905007177, 0.28847285171761333)
    (23.146725698865385, 0.2743608459472184)
    (18.785861496168685, 0.2602991823971855)
};
\addlegendentry{YOLOv8n}

\addplot[color=red, mark=square*] coordinates {
    (100, 0.3458179499730365)
    (79.9163903065935, 0.31436299871404916)
    (66.2555118139315, 0.30236394795647276)
    (56.71374020429718, 0.29401885633722163)
    (49.9474115709316, 0.2807181951128212)
    (45.36283800252602, 0.2601888376683355)
    (42.198373599426844, 0.23425407794306152)
};
\addlegendentry{YOLOv10n}

\addplot[color=green!60!black, mark=triangle*] coordinates {
    (100, 0.3332294561852834)
    (75.54020299355656, 0.3179408559367415)
    (58.83180258123796, 0.2983265983651694)
    (47.15137265802651, 0.2955518800336915)
    (38.826474869027074, 0.28693960301278815)
    (33.167131876319864, 0.27296502709774073)
    (29.250182416233304, 0.2646221915330499)
};
\addlegendentry{YOLOv11n}

\end{axis}
\end{tikzpicture}

\begin{tikzpicture}
\begin{axis}[
    width=9.4cm,
    height=7cm,
    xlabel={Parameters Retained (\%)},
    ylabel={MAC operations (in billions)},
    xmin=15, xmax=102,
    ymin=113360458, ymax=528707594,
    x dir=reverse,
    grid=major,
    grid style=dashed,
    legend pos=south west,
    legend style={font=\scriptsize},
    line width=1pt,
    tick label style={font=\scriptsize},
    label style={font=\scriptsize},
    enlargelimits=false,
]

\addplot[color=blue, mark=*] coordinates {
    (100, 498707594.0)
    (71.45755333852722, 373598638.0)
    (52.245654445115264, 285874232.0)
    (38.852159809354156, 226397640.0)
    (29.502516905007177, 181913578.0)
    (23.146725698865385, 152476436.0)
    (18.785861496168685, 133360458.0)
};
\addlegendentry{YOLOv8n}

\addplot[color=red, mark=square*] coordinates {
    (100, 509299924.0)
    (79.9163903065935, 415447235.0)
    (66.2555118139315, 350218043.0)
    (56.71374020429718, 307297179.0)
    (49.9474115709316, 275372699.0)
    (45.36283800252602, 254790004.0)
    (42.198373599426844, 242299953.0)
};
\addlegendentry{YOLOv10n}

\addplot[color=green!60!black, mark=triangle*] coordinates {
    (100, 390731978.0)
    (75.54020299355656, 301515032.0)
    (58.83180258123796, 239552180.0)
    (47.15137265802651, 198787316.0)
    (38.826474869027074, 168857528.0)
    (33.167131876319864, 149057363.0)
    (29.250182416233304, 136954412.0)
};
\addlegendentry{YOLOv11n}

\end{axis}
\end{tikzpicture}
\caption{Effect of pruning on a) accuracy and b) inference complexity, for YOLOv8n, YOLOv10n and YOLOv11n}
\label{MACs vs Pruning ratio}
\end{figure}


\begin{figure*}
    \centering
    \includegraphics[width=1\linewidth]{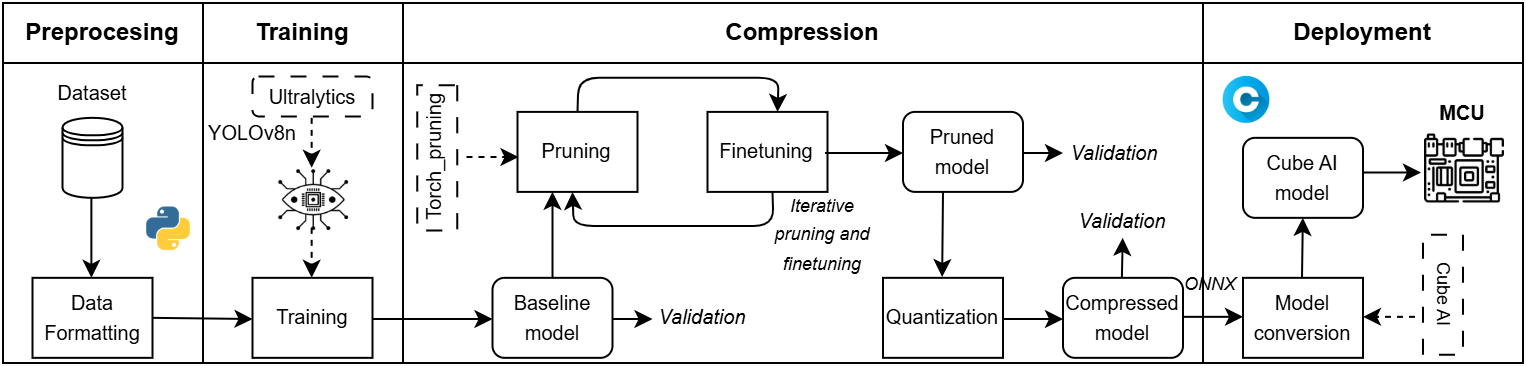}
    \caption{Diagram of the preprocessing, training, compression and deployment pipeline used.}
    \label{meth_diag}
\end{figure*}

The structured pruning implementation of this work is based on Torch-Pruning~\cite{fang2023depgraph}, an open-source pruning library that automatically traces and maintains model dependency graph. 
To determine which filters to prune, \textit{local pruning} or \textit{global pruning} can be used. With \textit{local pruning}, parameter importance is evaluated within the current layer while with \textit{global pruning}, a global ranking of all parameters is performed. The importance evaluation is commonly performed using \textit{L2 norm}. The L2 norm of a weight vector $w = (w_1,...,w_n)$ measures its magnitude and is defined as:

\begin{equation}
 \|w\|_2 = \sqrt{\sum_{i=1}^{n} w_i^2}
\end{equation}




To reach target sparsity, an iterative pruning schedule is selected to gradually prune the model, while allowing the network training steps to recover from any pruning-induced loss in accuracy. This approach is generally considered to achieve better accuracy than aggressive one-shot pruning, especially at higher pruning ratios~\cite{janusz2025shotvsiterativerethinking}, as it allows access to narrow, well-generalizing minima, which are typically ignored when using one-shot approaches~\cite{Tartaglione_2020}.
Given a target pruning ratio $r_{target}$ and a total number of steps $k$, the  pruning fraction per step is calculated using a progressive geometric decay schedule as:

\begin{equation}
\text{pruning per step} \;=\; 1 - \bigl(1 - {r_{target}}\bigr)^{\tfrac{1}{\text{k}}}
\end{equation}

Following this schedule, the network is pruned rapidly in the initial phase when the redundant connections are abundant while the number of weights to being pruned are gradually reduced following each step as there are fewer remaining in the network~\cite{zhu2017prunepruneexploringefficacy}.

As depicted on Figure~\ref{pruning_steps}, the network accuracy was evaluated for a pruning ratio of $70\%$, for $k \in \{1, \ldots, 12\}$. It can be seen that $k=6$ offers the best tradeoff between accuracy and training speed and was thus used for the rest of this work's experiments. The pruning process begun with the pre-processing and validation of the model to establish a baseline for accuracy.
Concatenation and detection layers were excluded due to their sensitivity to parameter reduction. At each iteration, the least important filters were pruned based on their L2 norm ranking, followed by fine-tuning the network for $30$ epochs to recover any accuracy loss. This cycle was repeated until the desired pruning ratio was reached or a predefined accuracy drop threshold of $20\%$ was met, in terms of \gls*{map}. The output of this process is a \textit{float32} model in \textit{onnx} format.


As can be seen in Figure~\ref{MACs vs Pruning ratio} (a), YOLOv8n demonstrated significantly higher robustness to pruning compared to YOLOv10n and YOLOv11n, with a notably lower performance degradation. Consequently YOLOv8n was selected as the baseline architecture for subsequent optimization experiments. Performance evaluation indicates that up to $60\%$ of the network's parameters can be effectively pruned with an \gls*{map} loss of $0.1$ compared to the unpruned model. A pruning ratio of $70\%$ pruning leads to a $0.15$ loss, which is still within acceptable limits for this scenario. However, more aggressive pruning results in a notable accuracy degradation, indicating a threshold beyond which the model no longer maintains sufficient representational capacity. Figure~\ref{MACs vs Pruning ratio} (b) depicts a near-linear correlation between the number of pruned parameters and the reduction in computational cost measured in \gls*{mac} operations.

Based on the evaluation steps above, a pruning ratio of $70\%$ was selected, leading to a total of $960$k parameters for the network.

\subsubsection{Quantization}
Quantization is one of the most widely adopted techniques for improving the efficiency of edge AI models and \gls*{ml} accelerators~\cite{verhelst2025keep}. It reduces the numerical precision of model parameters and activations, typically converting $32$-bit floating-point numbers into $8$-bit integers or lower. This results in a memory overhead reduction by a factor of $4\times$ for tensor storage, while the computational cost for matrix multiplications is reduced quadratically by a factor of $16$~\cite{nagel2021white}. Additionally, integer operations consume considerably less energy than their floating-point counterparts~\cite{horowitz20141}. 

The core idea behind quantization is to map real-valued numbers to a finite set of discrete levels using a linear transformation, also known as an affine mapping. This process is referred to as \textit{linear quantization}, which can be further categorized into \textit{asymmetric} and \textit{symmetric uniform quantization} schemes.
In asymmetric uniform quantization, the quantization process is defined as~\cite{jacob2018quantization}:

\begin{equation}
\begin{split}
\overline{x} = clamp \left( \lfloor \frac{x}{s} \rceil - \lfloor z \rceil, 0, 2^b-1 \right) \\ s = \frac{\theta_{max}-\theta_{min}}{2^b-1}\\
z = \frac{\theta_{min}}{s}
\end{split}
\end{equation}

where $x$ is the original real value input, $\overline{x}$ is the quantized integer output, $[\theta_{min},\theta_{max}]$ is the quantization range and $b$ is the target bit width. The clamp function ensures that the output is within the target range and is defined as:

\begin{gather}
clamp (x; a, c) =
\begin{cases}
 a, & x < a\\
 x, & a \le x \le c\\
 c, & x > c
\end{cases}
\end{gather}

Scale $s$, typically stored in a float $32$-bit format, determines the quantization granularity, while zero-point $z$ (an integer) shifts the quantized range to best approximate $0$ in the original domain. To recover an approximate real value $\hat{x}$ from the quantized integer, a de-quantization step is performed:

\begin{equation}
\hat{x} = s(\overline{x} +\lfloor z \rceil)     
\end{equation}

There are different approaches and design decisions that revolve around the selection of optimal quantization parameters $s$ and $z$ for weights and activations. The first decision concerns the choice of granularity~\cite{lin2025qservew4a8kv4quantizationcodesign}: \textit{Per-tensor} quantization shares $s$ and $z$ across the entire tensor. In \textit{per-channel} quantization for weights or \textit{per-token} quantization for activations, $s$ and $z$ are shared within each row of tensor. Finally, \textit{per-group} quantization further reduces the degree of parameter sharing by using different $s$ and $z$ for every $g$ columns within each row, where $g$ is the group size.

The next step is the definition of the range estimation method, i.e. the strategy to define $\theta_{min}$ and $\theta_{max}$. 
A simple range estimation method is the \textit{min-max} strategy, which sets $\theta_{min}$ and $\theta_{max}$ to the minimum and maximum values of the input data to cover the whole dynamic range of the tensor. While this method ensures full range coverage, it is highly sensitive to outliers. 
An alternative method is to minimize the \textit{\gls*{mse}} between the original and quantized tensors, which yields tighter bounds, mitigating the impact of large outliers. The optimization problem of minimizing \gls*{mse} is frequently solved using grid search. In certain cases, where all tensor values are not equally important, it can be effective to minimize the \textit{cross-entropy loss} instead, which focuses on preserving classification accuracy by emphasizing the most relevant logits.

There are two primary quantization workflows. In \textit{\gls*{ptq}}, weights are quantized ahead of time and the per-activation dynamic range is computed at runtime. In \textit{static} PTQ, both weight and activation tensors are quantized ahead of time. Since the activation tensors are not known at that stage, a small calibration dataset is required, to estimate activation ranges.
\textit{Quantization-aware training (QAT)}
simulates quantization effects during training using fake quantization operators (quantizer blocks), allowing the network to learn robustness to quantization noise. QAT typically leads to higher accuracy and supports more aggressive quantization schemes. A common practice is to apply per-channel or symmetric quantizers for the weights and asymmetric quantizers for the activations~\cite{nagel2021white}.

\textit{Symmetric quantization} is a simplified version of the general asymmetric case, where the zero-point is fixed to $0$, eliminating thus the offset-related computational overhead during the accumulation operation~\cite{nagel2021white}. While less flexible, this approach is computationally cheaper and is frequently used for weight quantization.

For this work, the \textit{onnxruntime.quantization} library was used to apply \textit{static} \gls*{ptq}. A calibration dataset of $300$ samples was used to estimate activation ranges. Initially, the \textit{onnx} model obtained after pruning was pre-processed to prepare it for quantization. The quantization was performed using the Quantize-DeQuantize (QDQ) format, which inserts \textit{QuantizeLinear} and \textit{DeQuantizeLinear} blocks around tensors to simulate low-precision inference. The \textit{min-max} range estimation strategy was used to determine the $s$ and $z$ parameters. To preserve model performance, only the convolutional, multiplication and addition operators were quantized. An asymmetric uniform quantization scheme was used for activations, while a symmetric scheme was applied to weights. Additionally, per-channel quantization was enabled for weights to improve accuracy and reduce quantization-induced degradation.

\subsection{Deployment}
\label{sec:deployment}

For the conversion of the compressed \textit{onnx} model into optimized c code, the proprietary STM32Cube.AI toolchain was used, an extension library of STM32CubeMX by ST Microelectronics.
Table~\ref{tab:compression} lists the memory and processing requirements of the final compressed network, after applying the model compression techniques detailed in section~\ref{sec:model_compression} above, and compares them to the original YOLOv8n model and the \gls*{mcu} specifications.

\begin{table}[h]
\centering
\caption{Compression results}
\label{tab:compression}
\begin{tabular}{|l|ccc|}
\hline
\textbf{Metric} &  \textbf{\shortstack{Baseline \\ YOLOv8n}} & \textbf{\shortstack{Compressed \\ YOLOv8n}} & \textbf{\shortstack{HW \\ constraints}}\\
\hline
Model size (Flash) & $9.95$ MB & \textbf{850.97 KB} & $2$ MB\\
RAM usage & $2.5$ MB & \textbf{677.30 KB} & $768$ KB\\
Parameters & $3200$M & \textbf{960k} & - \\
GMACs & $0.50$ & \textbf{0.143} & - \\
\hline
\end{tabular}
\end{table}


The compressed model was deployed on the NUCLEO-U575ZI-Q \gls*{soc}, integrating the STM32U575ZI \gls*{mcu}. The \gls*{mcu} features a $32$-bit ARM Cortex-M33 core, with $2$ MB of flash memory and $786$ KB of SRAM.
The core has a very low energy profile, with a $8.95\mu$A current consumption at Stop $2$ mode and $19.5\mu$A per MHz run mode at $3.3V$ according to the device datasheet.
For this deployment, the MCU was clocked at $160$ MHz via the \gls*{msi}. The internal low-power \gls*{rtc} of the MCU was used for time keeping purposes clocked through the \gls*{lsi} at $32$ kHz.
The integrated SMPS regulator was used for voltage and current control.
All unused GPIOs were set to reset state to minimize leakage currents.
The Hardware Abstraction Layer (HAL) library of the STM32CubeMX by ST Microelectronics was utilized to program the \gls*{soc}.
To ensure accurate current consumption measurement, the on-board LD5 LED was de-soldered as it was not controllable through software.

Upon power up, the board initializes all required peripherals and begins inference. After each inference cycle, the system enters Stop 2 mode via the \textit{HAL\_PWREx\_EnterSTOP2Mode()} HAL instruction combined with the Wait For Interrupt (WFI) flag. The system exits Stop 2 and resumes operation through a periodic interrupt, triggered every $30$ seconds by the RTC, upon which it re-initializes and repeats the inference process.


Figure~\ref{meth_diag} depicts a complete diagram of the proposed preprocessing, training, compression and deployment pipeline used in this work.


\section{Discussion}
\label{sec:evaluation}

This section presents an evaluation of the proposed compressed network deployed on the STM32U575ZI \gls*{mcu}. The CropAndWeed dataset was used for benchmarking, with an $80$ - $20$ training and validation split. Performance metrics include \gls*{map}50-95, the number of model parameters (M), latency per inference (ms), frame rate (\gls*{fps}) and energy consumption per inference.

\begin{table}[h]
\caption{Comparison of the proposed network with existing YOLO implementations for weeds detection}\label{yolo_results_weed}
\centering
\scalebox{0.78}{
\begin{tabular}{ |p{1cm} p{1.4cm} p{1.4cm} c p{1.1cm} p{1.2cm} c|}
\hline 
\textbf{Work} & \textbf{Detector} & \textbf{Device} &\textbf{mAP50} &\textbf{mAP50-95} & \textbf{Params (M)} & \textbf{Input size}\\
\hline 
\multirow{2}{*}{\cite{wang2025lightweight}} & YOLO-Weed Nano & \multirow{2}{*}{RTX 3090} & \multirow{2}{*}{$0.931$} & \multirow{2}{*}{-} & \multirow{2}{*}{$1.09$} & \multirow{2}{*}{$640\times640$} \\
\multirow{2}{*}{\cite{khater2025ecoweednet}} & \multirow{2}{*}{EcoWeedNet} & RTX 3080 Ti & \multirow{2}{*}{$0.952$} & \multirow{2}{*}{$0.889$} & \multirow{2}{*}{$2.78$} & \multirow{2}{*}{$640\times640$}\\
\multirow{2}{*}{\cite{fan2024yolo}} & YOLO-WDNet & Jetson AGX Xavier  & \multirow{2}{*}{$0.978$} & \multirow{2}{*}{-} & \multirow{2}{*}{$0.9$}& \multirow{2}{*}{$640\times640$}\\
\textbf{This work} & \multirow{2}{*}{\textbf{YOLOv8n}} & \multirow{2}{*}{\textbf{STM32U575ZI}}  & \multirow{2}{*}{\textbf{0.517}} & \multirow{2}{*}{\textbf{0.403}} &  \multirow{2}{*}{\textbf{0.96}} & \multirow{2}{*}{\textbf{224 x 224}} \\
\hline 
\end{tabular}}
\end{table}

\subsection{Model Performance Analysis}

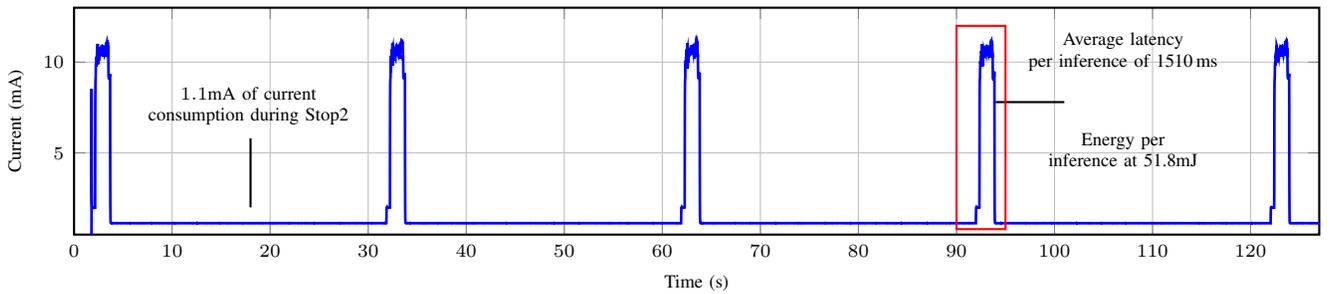
\begin{figure*}[ht]
    \centering
    \begin{tikzpicture}
        \begin{axis}[
    width=\linewidth,
    height=4.6cm,
            xlabel={Time (s)},
            ylabel={Current (mA)},
             ytick={0,5,10,15},              
    ymin=0.5, ymax=13,                
    scaled y ticks=false,           
            xmin=0, xmax=127,
            grid=both,
            thick,
            samples=100,
                tick label style={font=\scriptsize},
    label style={font=\scriptsize},
            every axis plot/.append style={line width=1pt},
        ]
        \addplot[
            color=blue,
            mark=none,
        ] table [
            x expr=\thisrowno{0}/1000,
            y expr=\thisrowno{1}/1000,      
            col sep=comma,
            skip first n=2,
        ] {ppk-20250815T123943.csv};

\node[anchor=south east, font=\scriptsize, align=center] at (axis cs:29,6) {$1.1$mA of current\\consumption during Stop2};
\draw[-, thick] (axis cs:18, 5.8) -- (axis cs:18, 2);  

\node[anchor=south, font=\scriptsize, align=center] at (axis cs:107,9) {Average latency\\per inference of 1510\,ms};
\draw[-, thick] (axis cs:94, 7.8) -- (axis cs:101, 7.8);
\node[anchor=south, font=\scriptsize, align=center] at (axis cs:107,3.7) {Energy per\\inference at 51.8mJ};
\draw[red, thick] (axis cs:90,0.8) rectangle (axis cs:95,12);

        \end{axis}
    \end{tikzpicture}
    \caption{Aggregated current draw of the system for $2$ mins.}
    \label{fig:current_plot}
\end{figure*}

Table~\ref{yolo_results_weed} presents the accuracy metrics of the proposed detection model when deployed for inference at the edge, as well as a comparison with recent state-of-the-art YOLO-based approaches for the detection of weeds. 
While prior works such as YOLO-Weed Nano \cite{wang2025lightweight}, EcoWeedNet \cite{khater2025ecoweednet}, and YOLO-WDNet \cite{fan2024yolo} achieve high detection accuracy, with \gls*{map}50 scores exceeding $93\%$, these models are designed for deployment on high-end platforms that lack stringent constraints on memory, power consumption, and compute resources. 
Indicatively, YOLO-Weed Nano has memory requirements for $2.4$MB for the network weights, while EcoWeedNet consists of triple the number of parameters as that of the proposed model.



Furthermore, the proposed model uses a significantly lower input resolution of $224\times224$ pixels to accommodate the limited memory specifications of the STM32U575ZI \gls*{mcu}. This contrasts with the $640\times640$ resolution used in most state-of-the-art works. Lower input resolution severely reduces spatial detail and limits the model’s ability to detect small or partially occluded objects, leading to a significant drop in detection accuracy.

Despite these limitations, the compressed model achieved a respectable \gls*{map}50 of $0.517$ and an \gls*{map}50-95 of $0.403$, which are sufficient for the detection of weeds in both open-air fields and greenhouses, where low-power operation outweighs the need for state-of-the-art accuracy.

\subsection{Energy Requirements} 

The current consumption of the \gls*{soc} was measured using Nordic Semiconductor Power Profiler Kit II (PPK2). PPK2 is a standalone unit supporting measurements from $200$nA up to $1$A, with a resolution between $100$nA to $1$mA depending on the range, for a supply voltage of $0.8$V - $5$V VCC. The \gls*{mcu} of this work was powered through the PPK2 using a $3.3$V supply voltage through a dedicated $3$V$3$ pin. 

\begin{table}[h]
\caption{Comparison of the proposed YOLOv8n network with existing low-power YOLO implementations}\label{yolo_results_stm}
 \centering
 \scalebox{0.8}{
    \begin{NiceTabular}[]{ | m[c]{8mm} m[l]{12mm} m[l]{15mm} m[c]{14mm} m[c]{10mm} m[c]{7mm} m[c]{7mm}|} 
    \hline 
        \textbf{Work} & \textbf{Detector} & \textbf{Device} &\textbf{Params (k)} & \textbf{Energy} & \textbf{Latency (ms)} & \textbf{Frame rate (fps)}\\ 
        \hline 
        \cite{moosmann2023tinyissimoyolo} & Tinyissimo YOLO & STM32H7A3, STM32L4R9, Apollo4b, MAX78000 &  $422$ & $41.8$mJ, $102$mJ, $6.08$mJ, $0.19$mJ  & $359$, $996$, $540$, $5.5$ & $2.79$, $1$, $1.85$, $181.82$\\
        \cite{boyle2024dsort} & Tinyissimo YOLO-based & GAP9 &  - &  $31 \mu$J & $16.2$ & $61.3$ \\
        \cite{humes2023squeezed} & Squeezed Edge YOLO & GAP8 & $931$ & $70.3$mJ & $130$ & $7.69$ \\
        \textbf{This work} & \textbf{YOLOv8n} & \textbf{STM32U575ZI} & \textbf{960}  & \textbf{51.8mJ} & \textbf{1510} & \textbf{0.66}\\

        \hline 
    \end{NiceTabular}
    }
\end{table}

Figure \ref{fig:current_plot} presents the system's current draw over a $2$-minute operation window.
During active inference, the system operates continuously, achieving an average latency of $1510$ms per inference, which corresponds to a frame rate of approximately $0.66$ \gls*{fps}. The average current measured over a complete inference cycle was $10.4$mA. The energy required for a single inference, given a supply voltage of $3.3$V, is then computed as:

\begin{equation}
\begin{split}
    E &= V \times I \times t \\&= 3.3 \times 10.4 \times 10^{-3} \times 1.510 \\&= 51.8mJ
\end{split}
\end{equation}

In low-power Stop $2$ mode, the \gls*{mcu} operates with an average current draw of approximately $1.1$mA for $30$ seconds, until re-activated through an RTC interrupt. Entering deep sleep between inference cycles significantly reduces total system energy consumption, enabling longer operation.
Using a $3.7$V Li-Ion battery with $7000$mAh ($25.9$Wh) the system can operate approximately for $189$ days of continuous operation, under ideal assumptions, without any human intervention.

A comparison of the proposed model of this work with compressed YOLO models from the literature for low-power embedded systems is given in Table~\ref{yolo_results_stm}. As can be seen from the Table, the proposed model of this work contains a significantly larger number of parameters, contributing to improved representational power. Despite the increased complexity, our system consumes $51.8$mJ per inference, remaining competitive in terms of energy. However, this comes at the cost of higher latency at $1510$ms, resulting in a lower throughput of $0.66$fps.

\section{Conclusions}
\label{sec:conclusions}

This work presents an optimized low-power computer vision pipeline for the detection of weeds in digital agriculture, leveraging structured pruning, post-training quantization, and input resolution scaling to compress the YOLOv8n detector for efficient deployment on the STM32U575ZI \gls*{mcu}. 

The compressed model achieves a substantial reduction in model size, from $9.95$ MB to $850.97$ KB, and a memory footprint that fits within the $2$ MB flash and $768$ KB SRAM constraints of the target hardware. Despite these aggressive optimizations, the system retains a respectable detection performance with an \gls*{map}50 of $0.517$ and \gls*{map}50-95 of $0.403$ on the CropAndWeed dataset, which is acceptable for many real-world agricultural settings.

In terms of energy and latency, the proposed system achieves inference at $51.8$mJ in active mode and a frame rate of 0.66 fps, making it suitable for periodic monitoring tasks in power-constrained edge deployments in both open-air fields and greenhouses. 

Future work will explore the integration of hardware accelerators, the use of additional network compression techniques including low rank approximation and transfer learning, as well as the use of Neural Architecture Search to further enhance model energy consumption while maintaining detection accuracy.


\textit{The authors commit to provide full access to the project source code after publication of the manuscript}.

\ifCLASSOPTIONcaptionsoff
  \newpage
\fi



\bibliographystyle{IEEEtran}
\bibliography{bibtex/bib/IEEEexample}
%



%

\begin{IEEEbiography}[{\includegraphics[width=1in,height=1.25in,clip,keepaspectratio]{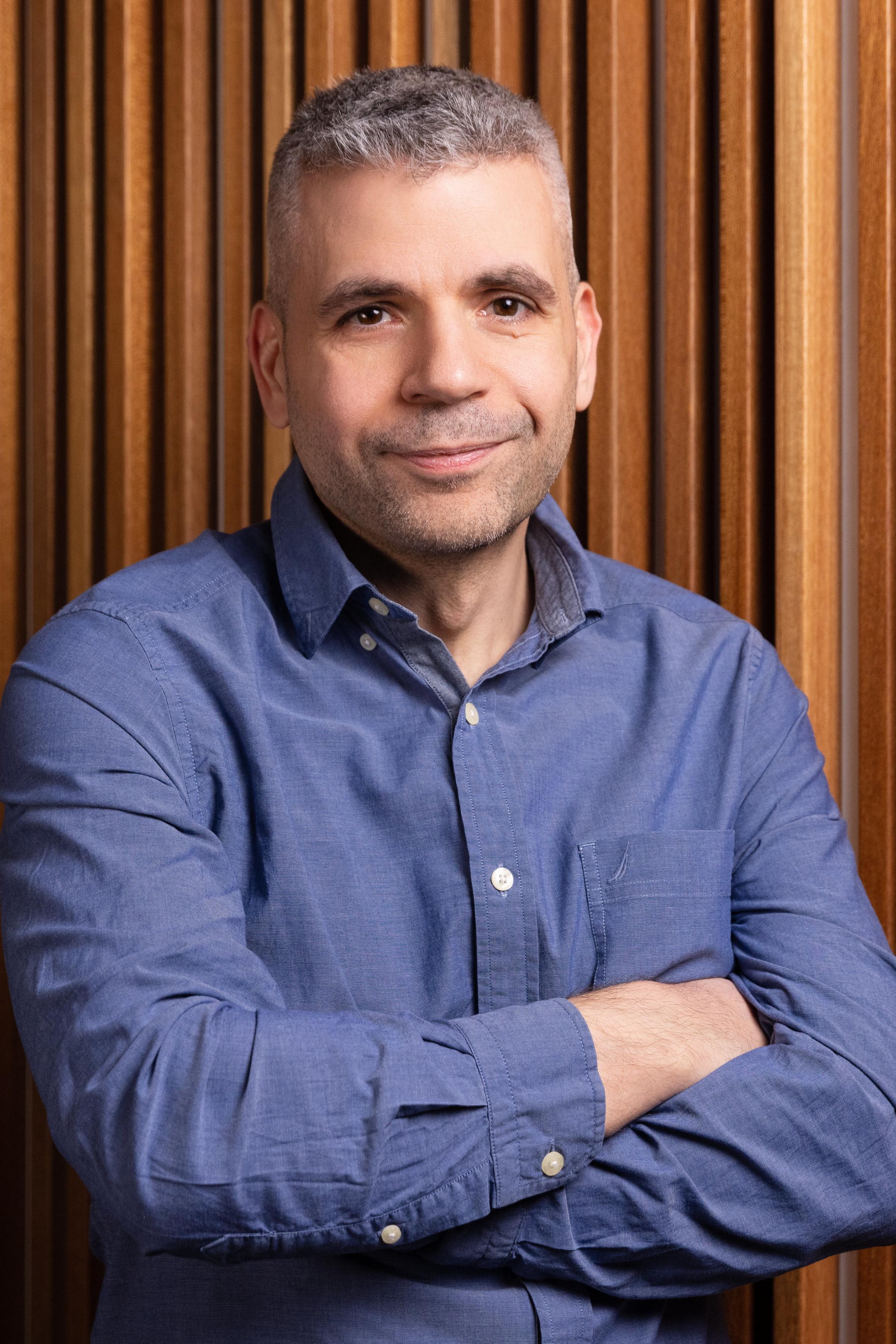}}]{Charalampos S. Kouzinopoulos}
is an Assistant Professor of the Internet of Things and Coordinator of the Computer Systems Research Area in the Department of Advanced Computing Sciences of the Faculty of Science and Engineering at Maastricht University. He was a Senior Research Fellow at the ALICE experiment of CERN and a Senior Researcher at CERTH/ITI.

His research interests lie in hardware design and integration, in software-hardware co-design and low-power edge AI, in computer architecture, parallel and distributed computing and in GPGPU computing.
\end{IEEEbiography}

\begin{IEEEbiography}[{\includegraphics[width=1in,height=1.25in,clip,keepaspectratio]{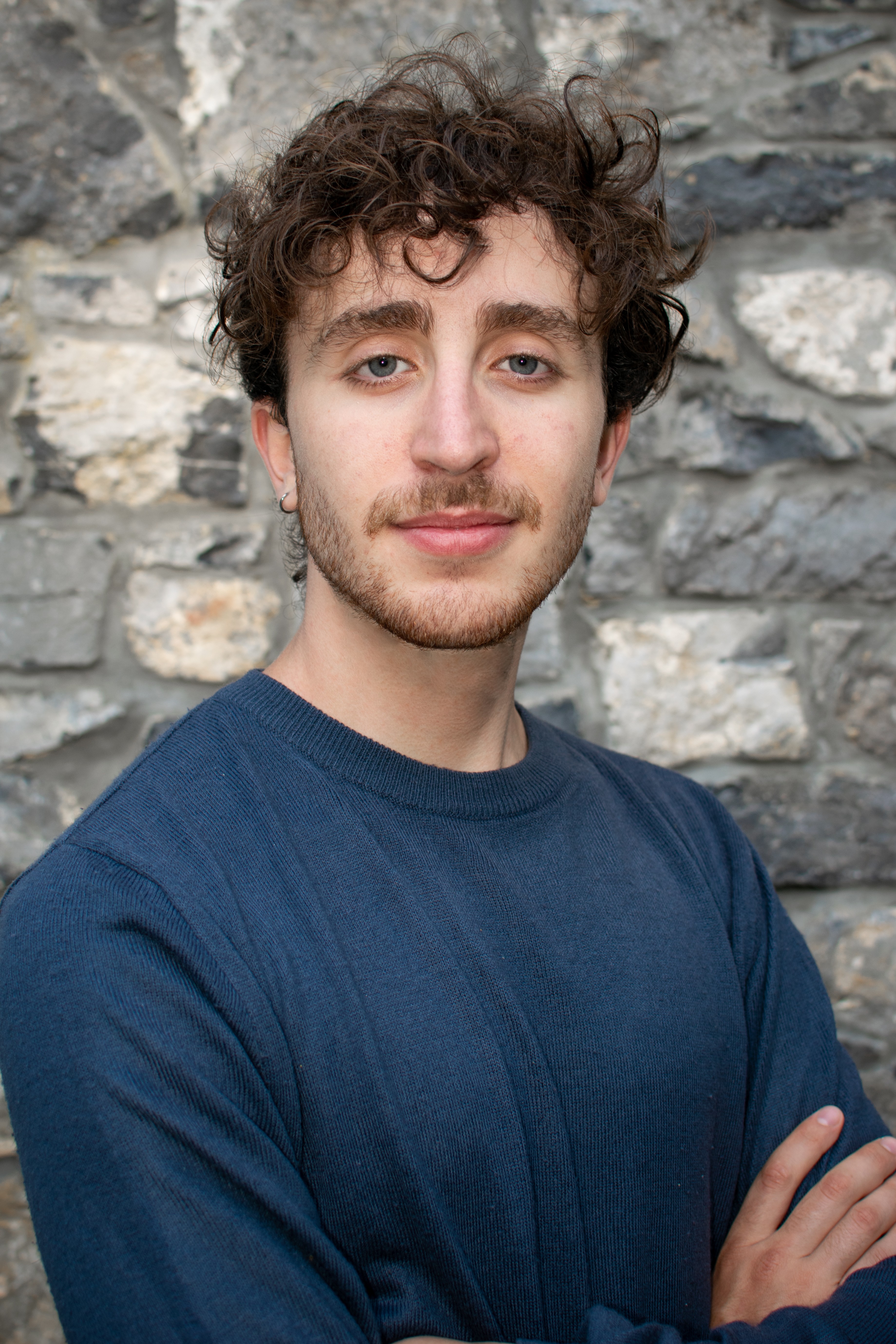}}]{Yuri Manna}
holds a BSc in Data Science and Artificial Intelligence from Maastricht University. His research focuses on AI systems, embedded and edge computing, and agricultural applications of artificial intelligence.
\end{IEEEbiography}





\end{document}